# Obstacle Avoidance for Autonomous Mobile Robots Based on Mapping Method

Anh-Tu Nguyen[1], Cong-Thanh Vu[2,*]

1 Faculty on Mechanical Engineering, Hanoi University of Industry, Hanoi, Vietnam

2 Faculty on Mechanical Engineering, University of Economics - Technology for Industries, Hanoi, Vietnam

*vcthanh@uneti.edu.vn

**Abstract.** In recent years, the mobile robot has been considerable attention to researchers for its application in various environments. For a mobile robot navigating its way from starting point to a goal point while traversing through deterrents, needs to recognize the obstacles and generate new trajectories to reach the destination. This paper presents an obstacle avoidance method for mobile robots using an open-source in robot operation system (ROS) combining with the dynamic window approach (DWA) algorithm. The experiment is carried out using a mobile robot in which the navigation data is based on data collecting by a laser scanner. The experimental results show that the robot can work well in environments containing static and dynamic obstacles.

## 1 Introduction

In recent decades, mobile robots have been applied in different fields of daily life. The mobile robot becomes more intelligent and can work autonomously in dynamic environments [1-2]. The researches in robot development can be classified as: localization, path planning, avoidance obstacle, and motion control [3]. Obstacle avoidance is a task in local path planning and this ensures the safety of humans and robots. To avoid collisions during motion, the mobile robot must detect boundaries of obstacles, create new trajectories, and calculate instant velocity and head angular [4].

There are numerous methods to implement avoidance collisions. One of the early approaches is the bug algorithm [5-6]. In this algorithm, the obstacles are first detected, then the robot moves following the boundary of the object. This algorithm only can work in the static environment. Another method that should be mentioned here is the potential field [7] that can be known as global path planning. This method assumes the target and the obstacles as the valleys and hills in a highland region, corresponding to the lowest and highest value of potential gravity. The robot will avoid obstacles bypassing the repulsive field and look forward to the goal using the attractive field. However, this algorithm still can not solve all the defects of the bug algorithms and it does not work well in narrow passages. Besides, the bug algorithm is difficult to apply to real-time systems. Another collision avoidance method is the histogram-based method. This method divides the robot working area into angular fields that can be transformed into a polar histogram. In this histogram, the proximity of an obstacle is described by each sector, then the robot's orientation is determined based on a cost function. The algorithms based on this method are VFH (Vector Field Histogram) [8] and VFH+ (Vector Field Histogram Plus) [9]. These algorithms can process better with ambiguity in the measurements and take into account the robot kinematics limitation. However, this method doesn't work well in potential local minima, especially with ''U'' shaped obstacles. Susnea et. al. proposed an approach in which the maximum available free space of the robot is defined as a bubble [10]. The bubble size can be extended in any direction without obstacles. Besides, the shape and size of the bubble are also determined by robot geometry and sensor measurement. Fox et. al. introduced a dynamic window approach (DWA) [11]. This method defines three velocities space: the maximum velocities window; obstacle-free field; and the admissible velocities. In the intersection of the three windows, the DWA algorithm chooses translational and rotational velocities maximizing the objective function.

This paper addresses a collision-avoidance approach for mobile robots using a combination of a 2D Costmap package on ROS and DWA algorithm. The environment data and obstacle information are collected through a laser scanner. The proposed method is examined by operating a mobile robot in static and dynamic environments. The results indicate that the robot can generate new paths to avoid obstacles during its motion.

## 2 Obstacle avoidance method

Many obstacle avoidance methods have been proposed in the field of robotics and each algorithm differs in the way of avoiding static or dynamic obstacles. By using a collision avoidance algorithm allows the robot to reach its target without colliding with any obstacles that may exist in its path. The obstacle avoidance algorithm is integrated with a local path planning algorithm, thus it has been used to modify the direction and the velocity of the mobile robot based on obstacles detection in its path. If any obstacle is found on the mobile robot path, the old path will be replaced with a new one to avoid that obstacle. In the present study, a *2D-Costmap* package in ROS is used to take in obstacle data through a laser scanner. The obstacle information is automatically updated and stored as purely a two-dimensional interface, therefore it can only be made in columns. Then, this information is used as boundary constraints of the dynamic window approach (DWA) optimization problem.

The (DWA) generates the local path for obstacle avoidance. The search for the robot controlling information is performed in the workspace of velocities. An optimal function is applied to determine the velocity maximizing of the objective function. The dynamic constraints are created to reduce the search space.

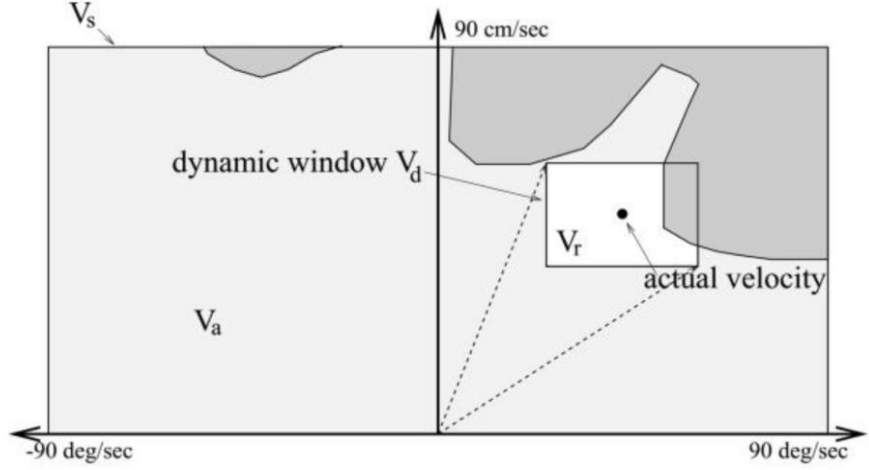

**Fig. 1.** The search space of DWA.

The search space $V_s$ of velocity is shown in Fig. 1, which allows the robot to stop before hitting the obstacle. The admissible velocities are defined as

$$V_a = \left\{(v,\omega) \,\middle|\, v \le \sqrt{2\text{dist}(v,\omega)\dot{v}} \wedge \omega \le \sqrt{2\text{dist}(v,\omega)\dot{\omega}}\right\} \quad (1)$$

Where: dist$(v,\omega)$ is the distance to the closest obstacle on the trajectory; $v$ and $\omega$ are translational and rotational velocities respectively; $\dot{v}$ and $\dot{\omega}$ are translational and rotational accelerations respectively.

Because the acceleration of the driven motor is limited, the velocity space is reduced to the dynamic window, $V_d$, that contains only reached velocities within a time step $\Delta t$.

$$V_d = \left\{(v,\omega) \,\middle|\, v \in [v_k - \dot{v}\Delta t, v_k + \dot{v}\Delta t] \wedge \omega \in [\omega_k - \dot{\omega}\Delta t, \omega_k + \dot{\omega}\Delta t]\right\} \quad (2)$$

The search space for the velocities within the dynamic window is defined as the intersection of the restriction areas:

$$V_r = V_s \cap V_a \cap V_d \quad (3)$$

After having the search space, the DWA evaluates each pair of $(v,\omega)$ in $V_r$ to obtain the optimal solution of the objective function:

$$G(v,\omega) = \alpha.\text{angle}(v,\omega) + \beta.\text{dist}(v,\omega) + \gamma.\text{vel}(v,\omega) \quad (4)$$

Where: angle$(v,\omega)$ is a measure of progress towards the goal position; vel$(v,\omega)$ is the forward speed of the robot and supports fast movements; and $\alpha$, $\beta$, $\gamma$ are constants of the objective function. The DWA algorithm is described in Fig. 2.

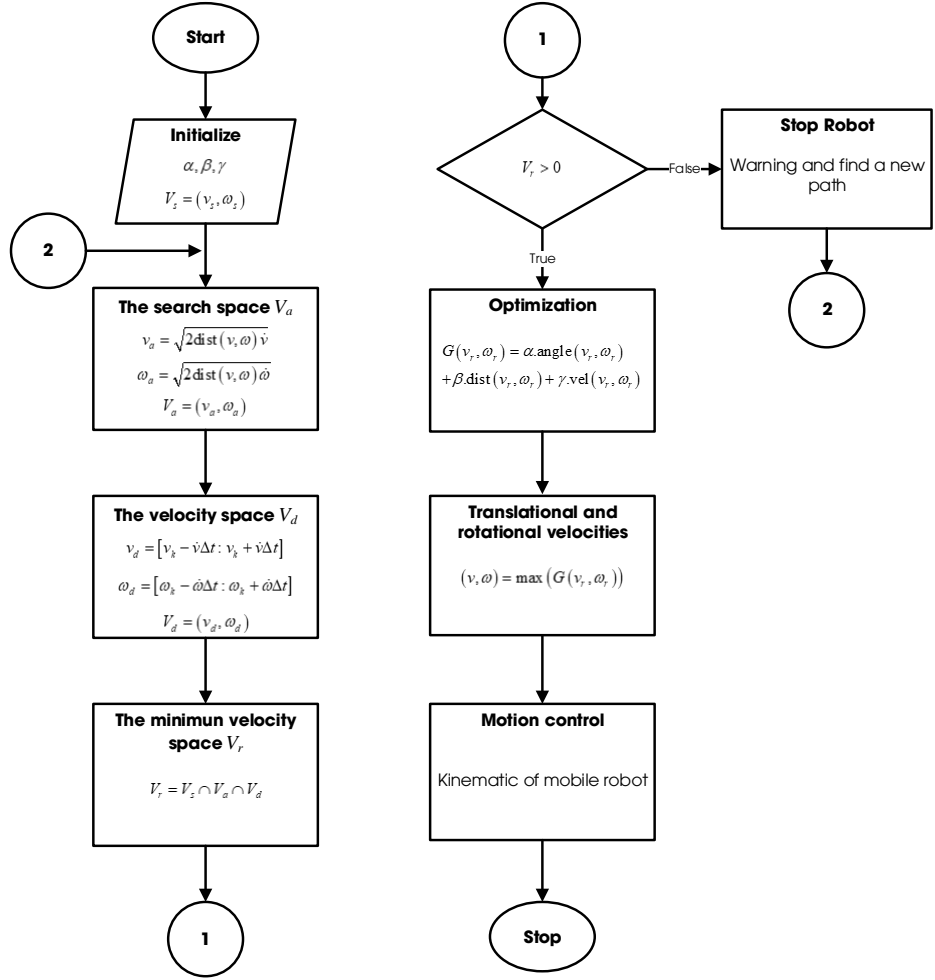

**Fig. 2.** The DWA algorithm.

## 3 Experimental robot

To evaluate the accuracy and stability of the proposed approach, an experimental robot was designed with two driving wheels and two castor freewheels (Fig. 3). The active wheels were driven using DC motors. The control system was developed based on a laptop integrating into the robot. A microcontroller, STM32F407VET6, received control signals from the laptop and transferred them to drive motors. The NAV245 was used to collect the environment information which would be applied for the localization and obstacle avoidance activity. Two encoder sensors were attached to the driven wheels, these sensors provided feedback signals for controlling motor speed.

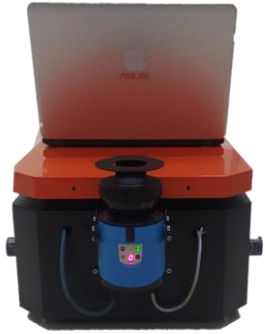 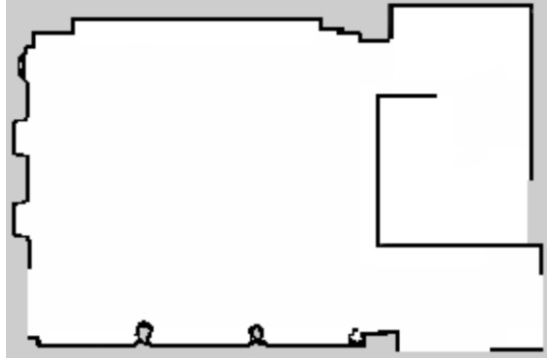

**Fig. 3.** Experiment robot.  **Fig. 4.** Global map of the environment.

The control program was developed using the ROS platform, in which a Hector Slam [12] package was applied for real-time map building (Fig. 4). To estimate the position of the robot in the environment, the adaptive Monte Carlo method (AMCL) was used in the localization system. This technique is also known as particle filter localization. The position and orientation data representing the robot's pose are re-sampled every time step for the robot localization (Fig. 5). Besides, the AMCL also helps recognize if any obstacle appears in the working area. After that, the DWA algorithm will calculate transitional and rotational velocities to control the robot the motion following the desired trajectories.

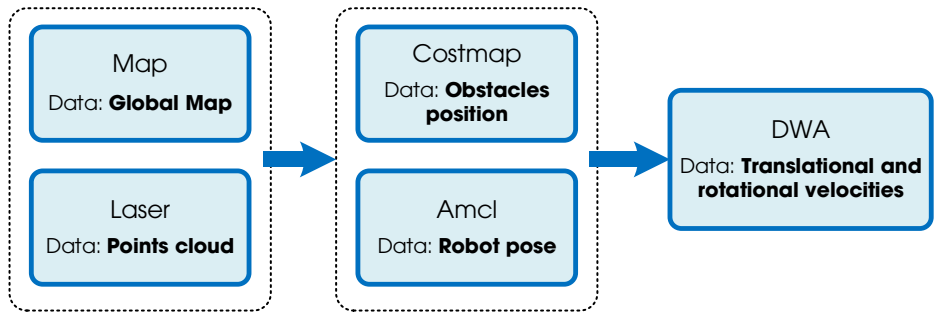

**Fig. 5.** The process schematic of the system.

## 4  Result and discussion

Based on the DWA to obstacle avoidance, an experimental robot operated in an indoor environment. The robot was programmed to automatically generate the trajectory to move from the start point to the goal point. The obstacles were designed with different boundaries (circle, rectangle), sizes, and different states (statics and dynamics) (Fig. 6).

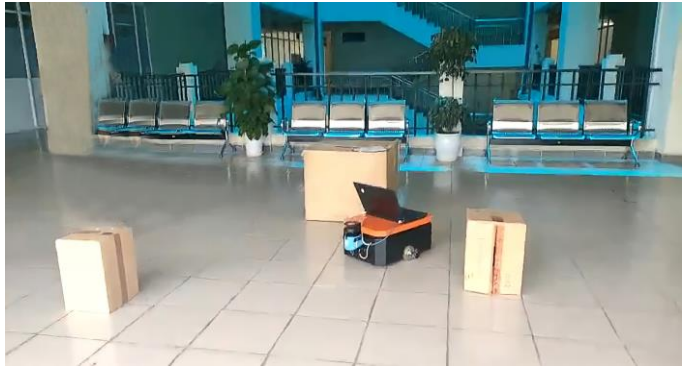

**Fig. 6.** Experiment environment.

To examine the proposed method, the obstacles were arranged at different positions in the working area, which required the robot to generate highly curvy paths to reach the goal point. Furthermore, the robot must determine instant velocity and head angular to avoid collision during its motion. The parameters of optimization function, $\alpha, \beta, \gamma,$ were chosen as in Table 1. In case 1, all obstacles were fixed in the working area (Fig. 7). It could be seen that the robot velocity increased rapidly to reach the maximal value of 0.6 m/s after 2 s, this value was remained for about 2.5 s before decreasing and fluctuating around 0.4 m/s. Then it decreased to finish the motion at the goal point (Fig. 8). The decrease in velocity is explained because the robot has to change its direction to avoid obstacles.

**Table 1.** The parameters of the DWA function.

| Scenario | $\alpha$ | $\beta$ | $\gamma$ |
|---|---|---|---|
| Case 1 | 0.85 | 0.15 | 0.1 |
| Case 2 | 1.0 | 0.1 | 0.5 |
| Case 3 | 1.0 | 0.1 | 0.1 |

In case 2, the experiment arrangement was shown in Fig. 9, only a dynamic obstacle (grey) was presented in the environment. This obstacle moved in an inverse direction to the robot. It is realized that the robot moved at maximal speed in almost the time, there was only a decrease in velocity when the robot came close to the obstacle. After passing the obstacle, the robot recovered its direction and maximal velocity before finishing the motion (Fig. 10). In case 3, the environment involved two static obstacles (blue) and a dynamic obstacle (grey). The results show similar features of velocity and angular as that of case 2 (Fig. 11). However, the robot had to change its direction early because of the presence of static obstacles. Besides, after passing the dynamic obstacle, the robot could not recover the maximal velocity because of the short remaining distance (Fig. 12).

The experimental results show that the Costmap package incorporates with a Laser scanner in this paper has successfully recognized both static and dynamic obstacles to provide information about obstacle boundaries. The DWA algorithm helps calculate instant velocity, angular velocity, and robot direction, which allows controlling the robot traversing through deterrents to get the goal point.

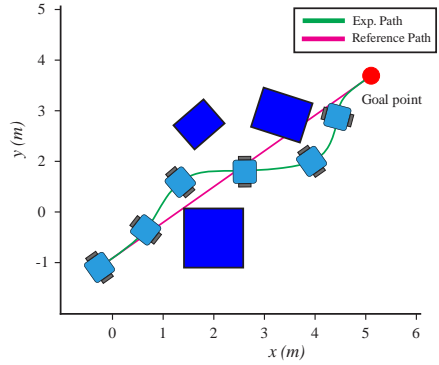
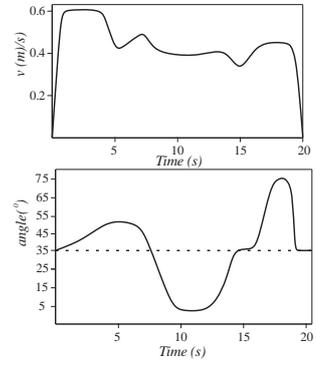

**Fig. 7.** The arrangement of obstacles in case 1.

**Fig. 8.** The translational velocity and robot angle of case 1.

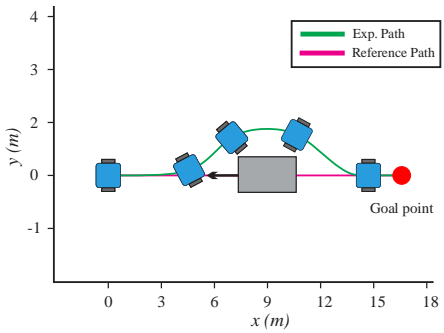
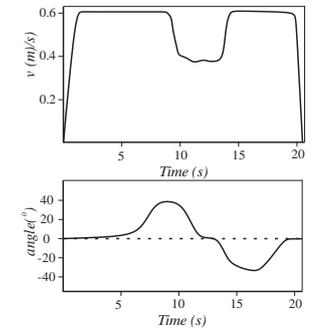

**Fig. 9.** The arrangement of obstacles in case 2.

**Fig. 10.** The translational velocity and robot angle of case 2.

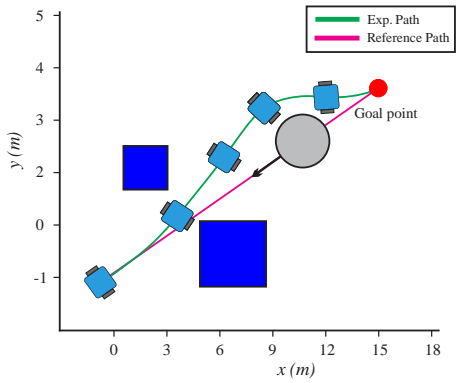
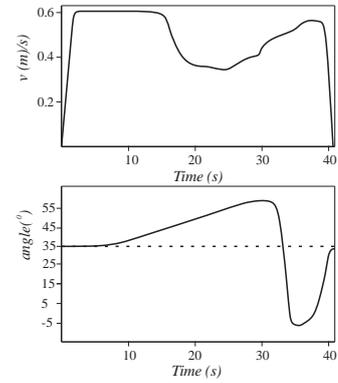

**Fig. 11.** The arrangement of obstacles in case 3.

**Fig. 12.** The translational velocity and robot angle of case 3.

## 5   Conclusion

The present paper introduces a collision-avoidance approach for mobile robots using a combination of different techniques. The information about the working environment is collected using a laser scanner for map building and obstacle detection. A combination of a 2D Costmap package on ROS and DWA algorithm has been used to calculate the translational and rotational velocities in the case of any obstacle occurring on the robot's motion path. Three experimental scenarios are carried out to examine the operation of the system. The robot has successfully passed the working area involving static and dynamic obstacles to reach the goal point. This allows extending the proposed approach to further study of mobile robot applications and to avoid dynamic moving obstacles in flexible environments.